\documentclass[english,utf8x]{article-hermes}

\usepackage{amsmath}
\usepackage{url}
\usepackage{booktabs}
\usepackage{graphicx}
\usepackage{pifont} 

\journal{TAL. Volume 57 -- n°3/2016}{1}{25}

\title[Sentence selection for exercises]{Candidate sentence selection for language learning exercises: from a comprehensive framework to an empirical evaluation}

\author{Ildikó Pilán \andauthor Elena Volodina \andauthor Lars Borin} 
\address{
Språkbanken, University of Gothenburg, \\ 
Box 100, 40530 Gothenburg, Sweden\\
\{ildiko.pilan, elena.volodina, lars.borin\}@gu.se\\
}

\abstract{We present a framework and its implementation relying on Natural Language Processing methods, which aims at the identification of exercise item candidates from corpora. The hybrid system combining heuristics and machine learning methods includes a number of relevant selection criteria. We focus on two fundamental aspects: linguistic complexity and the dependence of the extracted sentences on their original context. Previous work on exercise generation addressed these two criteria only to a limited extent, and a refined overall candidate sentence selection framework appears also to be lacking. In addition to a detailed description of the system, we present the results of an empirical evaluation conducted with language teachers and learners which indicate the usefulness of the system for educational purposes. We have integrated our system into a freely available online learning platform.
}


\motscles{
génération d'exercices,
concordances, 
évaluation orientée vers l'utilisateur.
}

\resume{
Nous proposons un système de traitement automatique de la langue ayant pour but
l'identification de phrases candidates tirées de corpus. Le système hybride allie une approche heuristique à des méthodes d'apprentissage automatisé et intègre un nombre de critères de sélection pertinents. Nous nous concentrons sur deux aspects fondamentaux : la complexité linguistique et la dépendance des phrases extraites envers leur contexte d'origine. Les travaux antérieurs en génération automatique d'exercices n'ont porté sur ces deux critères que de façon limitée, et un cadre fin de sélection de phrases candidates semble également faire défaut. 
En plus d'une description détaillée du système, cet article rapporte les résultats d'une évaluation empirique réalisée avec des enseignants de langues et des apprenants portant sur l'utilité du système à des fins éducatives. 
}

\begin{document}

\maketitlepage

\section{Introduction}
Several tasks related to foreign and second language (L2) learning can be partly or entirely automatized with the help of Natural Language Processing (NLP) tools. One such task is exercise generation, whose automation offers both self-directed learning opportunities and support for teaching professionals' practice. The pedagogical relevance and practical usefulness of such solutions, however, would need to be further improved before these systems can become widely used in language instruction. During our work, we aimed at maintaining a pedagogical angle, on the one hand, by incorporating statistical information from existing hand-written teaching materials into our selection criteria and, on the other hand, by evaluating the performance of our system with L2 teachers and learners.

Practice plays an important role in L2 learning  for the development of both receptive and productive skills \cite{dekeyser2007practice}. Corpora as potential practice material are readily available in large quantities, however, their use in L2 teaching has been both supported and opposed in previous years, \citeasnoun{o2007corpus} present an overview of this debate. Corpora offer a large amount of diverse examples at a low cost, and their use has been shown to have a positive effect on learners’ progress \cite{cobbthere,cresswell2007getting}. Moreover, corpora are evidence of real-life language use which, however, might be hard for learners to process \cite{kilgarriff2009corpora}. Non-authentic, teacher-constructed materials have also been subject to criticism. While this approach benefits from teachers' expert knowledge, these materials are “based on intuition about how we use language, rather than actual evidence of use” \cite[p.~21]{o2007corpus}. We aim at bringing together intuition and evidence about language use by employing insights from coursebooks for selecting examples from real-life corpora (e.g.~news texts, novels). 

Recent years have seen a number of efforts in the NLP community to automatically generate exercise items (e.g.~\cite{arregik2011automatic,smith2010gap,sumita2005measuring}). Most of these, however, tend to neglect what criteria sentences should fulfil in order to be suitable as exercise items and, instead, build on either a predefined set of manually selected sentences, or require merely a certain linguistic pattern (e.g.~a particular word) to be present in the sentence (see Section~\ref{ssec:ex_item}). When selecting sentences from corpora, however, there are a number of additional aspects that sentences need to adhere to in order to be usable and understandable in isolation. These have been previously explored mostly in a lexicographic context \cite{kilgarriff2008gdex}, but they are also relevant for language teaching \cite{kilgarriff2009corpora}. Two fundamental questions in this respect are: (i) Can the sentence function in isolation, outside its larger textual context? (ii) Is the complexity of the linguistic content of the sentence suitable for the intended L2 learner(s)?
We will refer to the former as \emph{context independence} and to the latter as \emph{L2 complexity}. 

Language learners pass through different learning stages (levels) reflecting the development and improvement of their competences. A scale of such levels is \textit{CEFR}, the Common European Framework of Reference for Languages \cite{councilofeurope2001}. The CEFR defines proficiency levels on a six-point scale: A1 (beginner), A2 (elementary), B1 (intermediate), B2 (upper intermediate), C1 (advanced) and C2 (mastery). A subset of language learners’ competences are \textit{linguistic competences}, which include, among others, lexical, grammatical and semantic competences. When assessing L2 complexity, we concentrate on linguistic competences required for reading comprehension since these can be matched to linguistic elements observable in language samples written for learners at different CEFR levels.

Both context independence and L2 complexity emerged as a main reason for discarding candidate sentences in previous evaluations \cite{arregik2011automatic,Pilan-Ildiko2013-9}, but thorough methods targeting these aspects have not been proposed up to date to our knowledge. 
Our approach, building on previous attempts at selecting sentences, contributes to previous research by offering a comprehensive set of criteria and by performing a more sophisticated selection in terms of the two fundamental aspects just mentioned, context independence and L2 complexity. We propose a hybrid system with both rule-based and machine learning driven components that encompasses a wide range of aspects. Incorporating rules makes the system customizable to users' needs and thus relevant for a wide range of application scenarios including vocabulary and grammatical exercises of different formats, as well as vocabulary examples. An evaluation with teachers and students indicates that our system identifies sentences that are, in general, of a suitable level of difficulty for learners. The algorithm is available to the general public free of charge both as a customizable sentence selection interface and as a web service. The development of automatically generated exercises using the selected sentences is also in progress.
Our target language is Swedish, a language for which the number of L2 learners has grown rapidly over recent years \cite{scb2016}. Although the current implementation is based on resources and tools for Swedish, the methods described can serve as an example for future implementations of exercise item candidate selection systems for other languages. 

This paper is structured as follows. In Section~\ref{sec:background}, we provide an overview of the related literature. Then, in Section~\ref{sec:implementation}, we describe our sentence selection framework in detail together with its implementation. Finally, in Section~\ref{sec:evaluation}, we present and discuss the results of a user-based evaluation of the system. 

\section{Related work}
\label{sec:background}

In this section, we provide an overview of the related literature which includes sentence selection strategies for both vocabulary examples and exercise items as well as studies on readability and CEFR level prediction.

\subsection{Sentence selection for vocabulary examples}
\label{ssec:context_indep}

GDEX, Good Dictionary Examples \cite{husak2010automatic,kilgarriff2008gdex} is an algorithm for selecting sentences from corpora for the purposes of illustrating the meaning and the usage of a lexical unit. It incorporates a number of linguistic criteria (e.g.~sentence length, vocabulary frequency, anaphoric pronouns) based on which example candidates are ranked. Some of these are related to context dependence (e.g.~incompleteness of sentences, presence of personal pronouns), but they are somewhat coarse-grained criteria without a focus on syntactic aspects.  

Besides English, the algorithm has also been successfully implemented for other languages. \citeasnoun{kosem2011gdex} and \citeasnoun{tiberius2015gdex} explore GDEX configurations for Slovene and Dutch respectively, aiming at identifying the optimal parameter settings for these languages for lexicographic projects.
 \citeasnoun{didakowski2012automatic} propose an example selection algorithm similar to GDEX for German. A fundamental difference of this method compared to the ranking mechanism of GDEX is having "hard criteria" which, if not met, result in sentences being excluded.
GDEX has also inspired a Swedish algorithm for sentence selection \cite{volodina2012semi} and it has been employed also for generating gap-fill exercises \cite{smith2010gap}. Furthermore, a number of machine learning approaches have been explored for these purposes in recent years \cite{geyken2015using,lemnitzer2015combining,ljubevsic2015predicting}. 
Example sentence selection for illustrating lexical items has also been addressed from a language teaching perspective by \citeasnoun{segler2007investigating}, where a set of selection criteria used by teachers was modelled with logistic regression. The main dimensions examined include syntactic complexity and similarity between the original context of a word and an example sentence.

\subsection{Sentence selection for exercise item generation}
\label{ssec:ex_item}

In a language-learning scenario, corpus example sentences can be useful both as exercise items and as vocabulary examples. Sentences used in exercises are also known as \textit{seed sentences} \cite{sumita2005measuring} or \textit{carrier sentences} \cite{smith2010gap} in the Intelligent, i.e.~NLP-enhanced, Computer-Assisted Language Learning (ICALL) literature.

Previous work on exercise item generation has taken into consideration a rather limited amount of aspects when selecting seed sentences. In some cases, sentences are only required to contain a lexical item or a linguistic pattern that constitutes the target of the exercise, but context dependence and L2 complexity are not explicitly addressed \cite{sumita2005measuring,Mitkov:2006:CEG:1133917.1133920,arregik2011automatic,wojatzki-melamud-zesch:2016:BEA11}. 
LanguageMuse \cite{burstein2012language}, a system supporting teachers in generating activities for learners of English, also belongs to this category. The sentences are selected from texts provided by teachers, the criteria of selection being the presence of a specific linguistic element that constitutes the target of the exercise: a lexical entity, a syntactic structure or a discourse relation.

Another alternative has been using dictionary examples as seed sentences, e.g.~from WordNet \cite{pino2009semi}. Such sentences are inherently context-independent, however, they impose some limitations on which linguistic aspects can be targeted in the exercises and they are not adjusted to finer-grained L2 learning levels. A system using GDEX for seed sentence selection is described in \citeasnoun{smith2010gap}, who underline the importance of the well-formedness of a sentence and determine a sufficient amount of context in terms of sentence length.
\citeasnoun{P16-4020} describe an ICALL system for fill-in-the-blanks preposition learning exercises, where seed sentences are checked for their lexical difficulty based on the level of the words according to a graded vocabulary lists. \citeasnoun{pilan-volodina-johansson:2014:W14-18} present and compare two algorithms for candidate sentence selection for Swedish, using both rule-based and machine learning methods. Context dependence, which has not been specifically targeted in their system, emerged as a key issue underlying suboptimal candidate sentences during an empirical evaluation. 

\subsection{Readability and proficiency level classification}
\label{ssec:readability}

The degree of complexity in the linguistic content of sentences and texts is one of the aspects underlying not only proficiency levels, but also readability. Readability measures typically classify texts into school grade levels or into a binary category of easy- vs. hard-to-read, but the term has also been used in the context of CEFR level classification, e.g.~\citeasnoun{xia-kochmar-briscoe:2016:BEA11}, \citeasnoun{franccois2012ai}.
In recent years a number of NLP-based readability models have been proposed not only for English   \cite{collins2004language,schwarm2005reading,graesser2011coh,vajjala2012improving,collinscomputational}, but also for other languages, e.g.~Italian \cite{dellorletta-montemagni-venturi:2011:SLPAT} and Swedish \cite{heimann2013see}. The linguistic features explored so far for this task include information, among others, from part-of-speech (POS) taggers and dependency parsers. Cognitively motivated features have also been proposed, for example, in the Coh-Metrix \cite{graesser2011coh}. Although the majority of previous work focuses primarily on text-level analysis, the concept of sentence-level readability has also emerged and attracted an increasing interest in recent years \cite{Pilan-Ildiko2013-9,Vajjala.Meurers-14-eacl,dell2014assessing}. 

The prediction of proficiency levels for L2 teaching materials using supervised machine learning methods has been explored for English \cite{heilman2007combining,huang2011robust,zhang2013feature,salesky-shen:2014:W14-18,xia-kochmar-briscoe:2016:BEA11}, French \cite{franccois2012ai}, Portuguese \cite{branco2014rolling}, Chinese \cite{sung2015leveling} and, without the use of NLP, for Dutch \cite{velleman2014online}. 

Readability formulas for the Swedish language have a long tradition. One of the most popular, easy-to-compute formulas is LIX (\emph{Läsbarhetsindex}, `Readability index') proposed by \citeasnoun{bjornsson1968lasbarhet}. This measure combines the average number of words per sentence in the text with the percentage of long words, i.e.~tokens consisting of more than six characters. Besides traditional formulas, supervised machine learning approaches have also been tested. A Swedish document-level readability model is described by \citeasnoun{heimann2013see} and \citeasnoun{falkenjack2013features}. \citeasnoun{pilan2015readable}, on the other hand, investigate L2 complexity for Swedish both at document and sentence level.

\section{HitEx: a candidate sentence selection framework and its implementation}
\label{sec:implementation}

In this section, we present our candidate sentence selection framework, HitEx (\textit{Hitta Exempel} ‘Find Exemples’) and its implementation. After an overall description, we introduce and motivate each selection criteria in Sections~\ref{ssec:trg_pattern} to \ref{ssec:lex_crit}.

\subsection{Overall description}
In Table~\ref{tab:criteria}, we show the selection criteria belonging to the proposed framework, grouped into broader categories. Each \emph{criterion} is used to scan a sentence for the presence (or the absence) of linguistic elements associated to its "goodness", i.e.~its suitability for the intended use. Most criteria target aspects that are negatively correlated to the goodness of a sentence. Certain selection criteria are associated with one (or more) numeric \textit{parameter(s)} that users can set, e.g.~the minimum and maximum number of tokens for the sentence length criterion.
The categories concerning the search term, well-formedness and context independence can be considered \emph{generic} criteria that are applicable for a number of different use cases, e.g.~different exercise types, vocabulary examples, while the rest of the criteria are more \emph{specific} for exercise item generation. 
In general, the sources that served as basis for these criteria include previous literature (Section \ref{sec:background}), L2 curricula and the qualitative results of previous user-based evaluations \cite{volodina2012semi,pilan-volodina-johansson:2014:W14-18}.

\begin{table} 
\centering
\begin{tabular}{cl|cl}
\toprule
{\bf Nr} & {\bf Criterion} & {\bf Nr} & {\bf Criterion} \\
\midrule
& \bf Search term & & \bf Additional structural criteria \\
1 & \it Absence of search term & 13 & Negative formulations \\
2 & Number of matches & 14 & \it Interrogative sentence \\
3 & \it Position of search term & 15 & \it Direct speech \\
& \bf Well-formedness & 16 & \it Answer to closed questions \\
4 & \it Dependency root & 17 & Modal verbs \\
5 & Ellipsis & 18 & Sentence length \\
6 & \it Incompleteness & & \bf Additional lexical criteria\\
7 & Non-lemmatized tokens & 19 & Difficult vocabulary \\
8 & Non-alphabetical tokens & 20 & Word frequency \\
& \bf Context independence & 21 & Out-of-vocabulary words \\
9 & \it Structural connective in isolation & 22 & Sensitive vocabulary\\
10 & Pronominal anaphora & 23 & Typicality \\
11 & Adverbial anaphora & 24 & Proper names \\
12 & \bf L2 complexity in CEFR level & 25 & Abbreviations\\
\bottomrule
\end{tabular} 
\caption{HitEx: a sentence selection framework.} 
\label{tab:criteria}
\end{table}

We implemented a \emph{hybrid system} which uses a combination of machine-learning methods for assessing L2 complexity and heuristic rules for all other criteria. The motivation behind using rules is, on the one hand, that certain linguistic elements are easily identifiable with such methods. On the other hand, a sufficient amount of training data encompassing the range of all possible exercise types would be extremely costly to create. Moreover, explicit rules make the sentence selection customizable to users' task-specific needs which increases the applicability of HitEx to a diverse set of situations. The criterion of L2 complexity has been implemented using machine learning methods since its assessment comprises multiple linguistic dimensions and data was available for approaching this sub-problem in a data-driven fashion.
A few selection criteria in our framework are re-implementations of those described by \citeasnoun{volodina2012semi} and \citeasnoun{pilan-volodina-johansson:2014:W14-18}. Major additions to previous work include: (i) L2 complexity assessment on a 5-level scale, vs.~a  previously available binary classification model by \citeasnoun{pilan-volodina-johansson:2014:W14-18}, (ii) typicality and (iii) the assessment of context dependence. Sensitive vocabulary filtering and the use of word frequencies from \textit{SVALex} \cite{FRANCOIS16.275}, a word list based on coursebook texts, are also novel aspects that we incorporated with the aim of making the sentence selection algorithm more pedagogically aware. 

Our implementation relies on a number of different NLP resources. Our system searches for sentence candidates via \emph{Korp} \cite{borin2012korp}, an online infrastructure providing access to a variety of (mostly) Swedish corpora. The concordance web service of Korp provides a list of corpus examples containing a certain user-specified search term, e.g.~an uninflected word, \textit{lemma} or a grammatical structure. Through Korp, a large variety of text genres are available such as novels, blogs, news and easy-to-read texts. All corpora are annotated at different linguistic levels, which include lemmatization, part-of-speech (POS) tagging and dependency parsing. HitEx assesses sentences based on these annotations as well as information from a number of Swedish lexical-semantic resources. A major lexical resource used is SALDO \cite{borin2013saldo} which is based on lexical-semantic closeness between word senses organized in a tree structure. 

As a first step in our sentence \emph{scoring algorithm}, for each candidate sentence $s$ \Pisymbol{psy}{206} $S$, we apply a linguistic criterion  $c$ \Pisymbol{psy}{206} $C$ to $s$ either as a \emph{filter} $f$ \Pisymbol{psy}{206} $F$ or as a \emph{ranker} $r$ \Pisymbol{psy}{206} $R$, that is $C = F \cup R$.
The application of each criterion $c_k$ to all the sentences, $c_k(S) = V_{c_k}$ is a set of criterion \emph{values} ($v_{c_k}$ \Pisymbol{psy}{206} $V_{c_k}$). $V_{c_k}=\{0,1\}$ when $c_k$ \Pisymbol{psy}{206} $F$ and $V_{c_k}$ \Pisymbol{psy}{206} $\mathbb{R}$ when $c_k$ \Pisymbol{psy}{206} $R$; for instance, when $c_k$ is the proper names criterion used as a ranker, $v_{{c_k}{s_i}}$ corresponds to the number of times a proper name appears in $s_i$ \Pisymbol{psy}{206} $S$.
If $s_i$ contains an undesired linguistic element associated to $c_k$ \Pisymbol{psy}{206} $F$, then $v_{{c_k}{s_i}}=1$, and $s_i$ is excluded from the ranking of suitable candidates. Further details about how we obtain $V_{c_k}$ are outlined in Sections~\ref{ssec:trg_pattern} to \ref{ssec:lex_crit}. 
Some criteria encode binary characteristics (e.g.~interrogative sentence), therefore, only $c_k$ \Pisymbol{psy}{206} $F$ holds for these. We present these in italics in Table~\ref{tab:criteria}.

To rank non-filtered sentences, we compute a \emph{goodness} score $G_{s_i}$ \Pisymbol{psy}{206} $\mathbb{N}$, which reflects the degree to which $s_i$ is a suitable candidate based on $R$. 
When $c_k$ \Pisymbol{psy}{206} $R$, $R=R^+ \cup R^-$, where $r^+$ \Pisymbol{psy}{206} $R^+$ is a \emph{positive ranker} for positively correlated properties with goodness, namely typicality and SVALex frequencies; and $r^-$ \Pisymbol{psy}{206} $R^-$ is a \emph{negative ranker} that includes all other criteria.
Based on $V_{c_k}$, we compute an intermediate (per-criterion) goodness score ($subG_{{c_k}{s_i}}$) for each $s_i$, by sorting $S$ based on $V_{c_k}$ and assigning the ranking position of $s_i$ according to $V_{c_k}$ to $subG_{{c_k}{s_i}}$. Consequently, the number of subscores will be equal to the number of $c_k$ \Pisymbol{psy}{206} $R$ selected. 
During this sorting, for $s_i$ \Pisymbol{psy}{206} $S$ and $s_j$ \Pisymbol{psy}{206} $S$, for $r^+$ $subG_{{c_k}{s_i}} \geq subG_{{c_k}{s_j}} \Leftrightarrow v_{{c_k}{s_i}} \geq v_{{c_k}{s_j}}$ holds, and for $r^-$ $subG_{{c_k}{s_i}} \geq subG_{{c_k}{s_j}} \Leftrightarrow v_{{c_k}{s_i}} \leq v_{{c_k}{s_j}}$ applies. In other words, we rank $S$ based on an ascending order of goodness if  $c_k$ \Pisymbol{psy}{206} $R^+$ and a descending order of badness if $c_k$ \Pisymbol{psy}{206} $R^-$. Therefore, more suitable candidates receive a higher $subG_{{c_k}{s_i}}$. For example, suppose $r_k^-$ is proper names and $s_i$ contains 2 proper names, while $s_j$ contains none; then $subG_{{c_k}{s_i}}=1$ and $subG_{{c_k}{s_j}}=2$. The score $G_{s_i}$ is then computed by summing all subscores, that is $G_{s_i}=\sum subG_{{c_k}{s_i}}$. Finally, candidate sentences are ordered in a decreasing order based on $G_{s_i}$. A weighting scheme similar to GDEX would be possible with the availability of data specific for the end use of the sentences from where to estimate these weights. At the current stage, all ranking criteria contribute equally to $G_{s_i}$.
Suboptimal sentences containing elements to filter can also be retained and ranked separately, if so wished, based on the amount of $F$ matched. 
The final results include, for each $s_i$ \Pisymbol{psy}{206} $S$, its summed overall score ($G_{s_i}$), its final rank and \emph{detailed information} per selection criteria, as the screenshot presenting the system's graphical user interface in Figure~\ref{img:gui} in Section~\ref{sec:platform} shows. In the following subsections, we present each criterion in detail, grouped into categories.

\subsection{Search term}
\label{ssec:trg_pattern}
A \emph{search term} corresponds to one (or more) linguistic element(s) that users would like the selected sentences to contain. It can be either a lexical element such as an inflected word or a lemma; or a grammatical pattern, e.g.~verbs in a certain tense followed by a noun. The presence of a search term is guaranteed through the mere use of the Korp concordance web service which only returns sentences containing the searched expression. In some application scenarios, repeated matches of the search term may be considered suboptimal \cite[p.~157]{kosem2011gdex}, therefore we include this aspect among our criteria. Similarly, there might be a preference for the \textit{position} of the search term in the sentence in some use cases such as dictionary examples \cite{kilgarriff2008gdex}. 

\subsection{Well-formedness}
Good candidate sentences from corpora should be structurally and lexically well-formed \cite{kilgarriff2008gdex,husak2010automatic}. We incorporate two criteria targeting the former aspect: one can check sentences for the presence of a dependency \emph{root}, and \emph{ellipsis}, i.e.~the lack of a subject or a finite verb (all verb forms except infinitive, supine and participle) inspired by \citeasnoun{volodina2012semi}. The completeness criterion checks the beginning and the end of a sentence for orthographic clues such as capital letters and punctuation, in a similar fashion to \citeasnoun{pilan:2016:BEA11}. A large amount of \emph{non-lemmatized tokens}, i.e.~tokens for which no matching dictionary form could be identified (in the SALDO lexicon in our case), are also preferably avoided \cite[p.~15]{husak2010automatic}. These are mostly cases of spelling or optical character recognition errors, foreign words, infrequent compounds, etc. A large portion of \emph{non-alphabetical tokens} could be e.g.~a sign of mark-up traces in web material, which has a negative influence on the L2 complexity and the usability of a sentence \cite[p.~15]{husak2010automatic}. Users can specify a constant as a threshold for these criteria to determine the allowed amount of non-lemmatized and non-alphabetical tokens in a sentence.

\subsection{Context independence}
\label{ssec:context_indep_crit}
Since sentences originally form part of coherent texts, a crucial aspect to take into consideration during selection is whether sentences would be meaningful also as a stand-alone unit without their original, larger context. The presence of linguistic elements responsible for connecting sentences at a syntactic or semantic level is therefore suboptimal \cite{kilgarriff2008gdex}. We incorporate a number of criteria for capturing this aspect which we described also in \citeasnoun{pilan:2016:BEA11}. 

Syntactic aspects include \emph{structural connectives}, i.e.~conjunctions and subjunctions \cite{webber2003anaphora}. Two concepts connected by structural connectives may appear in separate sentences which give rise to context dependence. Our system considers sentences with connectives in sentence-initial position context dependent unless the sentence consists of more than one clause. Connectives that are paired conjunctions are also allowed (e.g.~\textit{antingen ... eller} `either ... or').

\emph{Anaphoric expressions} referring to previously mentioned information are aspects related to the semantic dimension. Our pronominal anaphora criterion targets mentions of the third person singular pronouns \textit{den} ‘it’ (common gender) and \textit{det} ‘it’ (neuter gender) as well as the demonstrative pronouns (e.g.~\textit{denna} ‘this’, \textit{sådan} ‘such’ etc.). The non-anaphoric use of \textit{det} (e.g.~in clefts: \textit{It is carrots that they eat.}), however, is not counted here. Such cases can be distinguished based on the output of the dependency parser: these occurrences of \textit{det} are tagged as expletive (pleonastic). Pronouns followed by a relative clause introduced by \textit{som} `which' are also considered non-anaphoric. 

Under \textit{adverbial anaphora}, we count time and location adverbs that behave anaphorically (e.g.~\textit{då} ‘then’) \cite{webber2003anaphora}. Another group of adverbs relevant for anaphora are those expressing logical relations (e.g.~\textit{istället} ‘instead’), which are also referred to as \textit{discourse connectives} \cite{webber2003anaphora}. Based on
\citeasnoun{teleman1999svenska}, a list of anaphoric adverbs has been collected and sentences are checked for the occurrence of any of the listed items.

\subsection{L2 complexity}
\label{ssec:l2_complexity_crit}
The aspect of L2 complexity has been assessed with the help of a supervised machine learning algorithm based on a number of different linguistic dimensions. We used the CEFR level classifier for sentences that we previously described in \citeasnoun{pilan2015readable}. The source of the training data was single sentences from COCTAILL \cite{volodina22you}, a corpus of coursebook texts for L2 Swedish. Such single sentences occurred either in the form of lists or so-called \emph{language examples}, sentences exemplifying a lexical or a grammatical pattern. The feature set used for assessing L2 complexity is presented in Table~\ref{table:feature_set}. This set consists of five subgroups of features: count-based, lexical, morphological, syntactic, and semantic features.

\begin{table}[h]
\begin{center}
\begin{tabular}{ll|ll}
\toprule
\bf Name & \bf Type & \bf Name & \bf Type\\
\midrule
Sentence length & \textsc{Count} &  Modal V to V & \textsc{Morph}\\
Avg token length &  \textsc{Count} & Particle IS & \textsc{Morph}\\
Extra-long token &  \textsc{Count} & 3SG pronoun IS & \textsc{Morph}\\
Nr characters & \textsc{Count} & Punctuation IS & \textsc{Morph}\\
LIX &  \textsc{Count} & Subjunction IS & \textsc{Morph}\\
Bilog TTR & \textsc{Count} & PR to N & \textsc{Morph}\\
Square root TTR &  \textsc{Count} & PR to PP & \textsc{Morph}\\
\cline{1-2}
Avg \textsc{KELLY} log freq & \textsc{Lexical} & S-V IS & \textsc{Morph}\\
A1 lemma IS & \textsc{Lexical} & S-V to V & \textsc{Morph}\\
A2 lemma IS & \textsc{Lexical} & ADJ IS & \textsc{Morph}\\
B1 lemma IS & \textsc{Lexical} & ADJ variation & \textsc{Morph}\\
B2 lemma IS & \textsc{Lexical} & ADV IS & \textsc{Morph}\\
C1 lemma IS & \textsc{Lexical} & ADV variation & \textsc{Morph}\\
C2 lemma IS & \textsc{Lexical} & N IS & \textsc{Morph}\\
Difficult W IS & \textsc{Lexical} & N variation & \textsc{Morph}\\
Difficult N\&V IS & \textsc{Lexical} & V IS & \textsc{Morph}\\
OOV IS & \textsc{Lexical} & V variation & \textsc{Morph}\\
No lemma IS & \textsc{Lexical} & Function W IS & \textsc{Morph}\\
\cline{1-2}
Avg. DepArc length & \textsc{Syntactic} & Neuter N IS & \textsc{Morph} \\
DepArc Len $>$ 5 & \textsc{Syntactic} & CJ + SJ IS & \textsc{Morph}\\
Max length DepArc &  \textsc{Syntactic} & Past PC to V & \textsc{Morph}\\
Right DepArc Ratio & \textsc{Syntactic} & Present PC to V & \textsc{Morph} \\
Left DepArc Ratio & \textsc{Syntactic} & Past V to V & \textsc{Morph} \\
Modifier variation & \textsc{Syntactic} & Supine V to V & \textsc{Morph}\\
Pre-modifier IS & \textsc{Syntactic} & Present V to V & \textsc{Morph}\\
Post-modifier IS & \textsc{Syntactic} & Nominal ratio & \textsc{Morph}\\
Subordinate IS & \textsc{Syntactic} & N to V & \textsc{Morph}\\
Relative clause IS & \textsc{Syntactic} & Lex T to non-lex T & \textsc{Morph} \\
PP complement IS & \textsc{Syntactic} & Lex T to Nr T & \textsc{Morph}\\
\cline{1-2}
Avg senses per token & \textsc{Semantic} & Relative structure IS & \textsc{Morph} \\
N senses per N & \textsc{Semantic} \\
\bottomrule
\end{tabular}
\end{center}
\caption{\label{table:feature_set} The feature set for L2 complexity assessment.}
\end{table}

\textit{Count features} are based on the number of characters and tokens (\textit{T}), extra-long words being tokens longer than 13 characters. LIX, a traditional Swedish readability formula (see Section~\ref{sec:background}) combines the sum of the average number of words per sentence in the text and the percentage of tokens longer than six characters \cite{bjornsson1968lasbarhet}. Bi-logarithmic and a square root type-token ratio (TTR) related to vocabulary richness \cite{heimann2013see} are also computed. 

\textit{Lexical features} incorporate information from the KELLY list \cite{volodina2012introducing}, a word list with frequencies calculated from a corpus of web texts (thus completely independent of the sentences in the dataset). KELLY provides a suggested CEFR level per each listed lemma based on frequency bands.
For some feature values, \textit{incidence scores} (IS) normalized values per 1,000 tokens are computed, which reduces the influence of sentence length. Word forms or lemmas themselves are not used as features, the IS of their corresponding CEFR level is considered instead. 
\textit{Difficult} tokens are those that belong to levels above the overall CEFR level of the text. Moreover, we consider the IS of tokens not present in KELLY (\textit{OOV IS}), the IS of tokens for which the lemmatizer could not identify a corresponding lemma (\textit{No lemma IS}), as well as average KELLY log frequencies.

\textit{Morphological features} include both IS and variational scores, i.e.~the ratio of a category to the ratio of lexical tokens: nouns (N), verbs (V), adjectives (ADJ) and adverbs (ADV).
The IS of all lexical categories as well as the IS of punctuation, particles, sub- and conjunctions (SJ, CJ) are taken into consideration. In Swedish, a special group of verbs ending in -s are called \emph{s-verbs} (\textit{S-VB}). These can indicate either a reciprocal verb, a passive construction or a deponent verb (active in meaning but passive in form). Due to their morphological and semantic peculiarity, they are explicitly targeted in L2 grammar books \cite{formifocus}.
Nominal ratio \cite{hultman1977gymnasistsvenska} is another readability formula proposed for Swedish that corresponds to the ratio of nominal categories, i.e.~nouns, prepositions (PP) and participles to the ratio of verbal categories, namely pronouns (PR), adverbs, and verbs. Relative structures consist of relative adverbs, determiners, pronouns and possessives. 

\textit{Syntactic features} are based, among others, on the length (depth) and the direction of dependency arcs (\textit{DepArc}). These aspects are related to readers working memory load when processing sentences \cite{gibson1998linguistic}.
For similar reasons, we consider also relative clauses as well as pre- and post-modifiers, which include, for example, adjectives and prepositional phrases respectively. 

\textit{Semantic features} draw on information from the SALDO lexicon. We use the average number of senses per token and the average number of noun senses per noun. Polysemous words can be demanding for readers as they need to be disambiguated for a full understanding of the sentence \cite{graesser2011coh}.

\citeasnoun{pilan2015readable} utilizing the feature set described above report 63.4\% accuracy using a logistic regression classifier for the identification of CEFR levels with an exact match, and 92\% accuracy for classifications within a distance of one CEFR level.
Besides the features outlined above, also the lack of culture-specific knowledge can be a factor influencing L2 complexity, as well as learners' knowledge of other languages. We, however, do not address these dimensions in the current stage due to a lack of relevant data.

\subsection{Additional structural criteria}

Besides the aspects mentioned above, a number of additional structural criteria are available which proved to be relevant either based on previous evaluations \cite{volodina2012semi,Pilan-Ildiko2013-9} or evidence from coursebooks \cite{volodina22you}. One such aspect is \emph{negative wording} which is preferable to avoid in exercise items \cite{frey2005item}. All tokens with the dependency tag of negation adverbials fall under this criterion. Under the \emph{interrogative sentence} criterion, we handle direct questions ending with a question mark. To detect \emph{direct speech}, we have compiled a list of verbs denoting the act of speaking based on the Swedish FrameNet \cite{heppin2012rocky}. The list contains 107 verbs belonging to frames relevant to speaking (e.g.~\textit{viska} `whisper' from the \textit{Communication manner} frame). This is combined with POS tag patterns composed of a minor delimiter (e.g.~dash, comma) or a pairwise delimiter (e.g.~quotation marks), followed by a speaking verb (optionally combined with auxiliary verbs), followed by a pronoun or a proper name. Both questions and sentences containing direct speech tend to be less common as exercise items, incorporating these among our criteria allows users to avoid such sentences if so wished.

\emph{Answers to polar (or close-ended) questions} are rarely employed as exercise items and they were also negatively perceived in previous evaluations \cite{volodina2012semi,Pilan-Ildiko2013-9}. This aspect is also relevant to the dependence of a sentence on a wider context. The algorithm tries to capture sentences of this type based on POS patterns: sentence-initial adverbs and interjections (e.g.~\textit{ja} `yes', \textit{nej} `no') preceded and followed by minor delimiters where the initial delimiter is optional for interjections. \emph{Modal verbs} were identified based on a small set of verbs used typically (but not exclusively) as modal verbs (e.g.~\textit{kan} `can', `know') where the dependency relation tag indicating a verb group excludes the non-auxiliary use. \emph{Sentence length}, a criterion which is also part of GDEX, is measured as the number of tokens including punctuation in our system. 

\subsection{Additional lexical criteria}
\label{ssec:lex_crit}

HitEx includes also options for filtering and ranking sentences based on information from lexical resources for ensuring an explicit control of this crucial aspect \cite{segler2007investigating}. Sentences containing \textit{difficult words}, i.e.~words whose CEFR level is above the target CEFR level according to the KELLY list, can be penalized or filtered. Besides KELLY, we also integrated into our system information from the SVALex list based on word frequencies from coursebook texts. Sentences with words absent from SVALex or with words below the average frequency threshold for the target CEFR level are thus additional scoring criteria. Another criterion involves the presence of \textit{proper names} which, although undesirable for dictionary examples \cite{kilgarriff2008gdex}, may be familiar and easy to understand for L2 students \cite{segler2007investigating}. Both proper names and \textit{abbreviations} were counted based on the POS tagger output.

In a pedagogical setting, certain \textit{sensitive vocabulary items} and topics tend to be avoided by coursebook writers. These are also referred to as PARSNIPs, which stands for Politics, Alcohol, Religion, Sex, Narcotics, Isms\footnote{An ideology or practice, typically ending with the suffix \textit{-ism}, e.g.~\textit{anarchism}.} and Pork \cite{gray2010construction}. Some topics are perceived as taboos cross-culturally, such as swear words, while others may be more culture-bound.
We compiled a word list starting with an initial group of seed words from more generally undesirable domains (e.g.~swear words) collected from different lexical and collaborative online resources (e.g.~Wikipedia\footnote{\url{https://www.wikipedia.org}.}) complemented with a few manually added entries. Furthermore, we expanded this automatically with all child node senses from SALDO for terms which represent sensitive topics (e.g.~\textit{narkotika} `narcotics', \textit{svordom} `profanities', \textit{mörda} `murder', etc.) so that synonyms and hyperonyms would also be included. A few common English swear words that are frequently used in Swedish in an informal context (e.g.~blog texts) were also incorporated. The current list of 263 items is not exhaustive and can be expanded in the future. The implementation allows teaching professionals to make the pedagogical decision of tailoring the subset of topics to use to a specific group of learners during the sentence selection. 

\textit{Typicality} can be an indication of more or less easily recognizable meaning of concepts without a larger context \cite{barsalou1982context}. We assessed the typicality of sentences with the help of a co-occurrence measure: \emph{Lexicographers' Mutual Information} (LMI) score \cite{kilgarriff2004itri}. LMI measures the probability of two words co-occurring together in a corpus and it offers the advantage of balancing out the preference of the Mutual Information score for low-frequency words \cite{bordag2008comparison}. We used a web service offered by Korp for computing LMI scores based on Swedish corpora. As a first step in computing the LMI scores, we collected nouns and verbs in the KELLY and SVALex lists (removing duplicates), which resulted in a list of 12,484~items. Then using these, we estimated LMI scores for all noun-verb combinations (nouns being subjects or objects) as well as LMI for nouns and their attributes using Korp. Counts were based on 8 corpora  of different genres amounting to a total of 209,110,000 tokens. The resulting list of commonly co-occurring word pairs consisted of 99,112 entries. Only pairs with a threshold of LMI $\geq$ 50 were included. The typicality value of a candidate sentence corresponded to the sum of all LMI scores available in the compiled list for each noun and verb in the sentence. 

\subsection{Integration into an online platform}
\label{sec:platform}

\begin{figure}
\centering
\fbox{\includegraphics[width=0.90\textwidth]{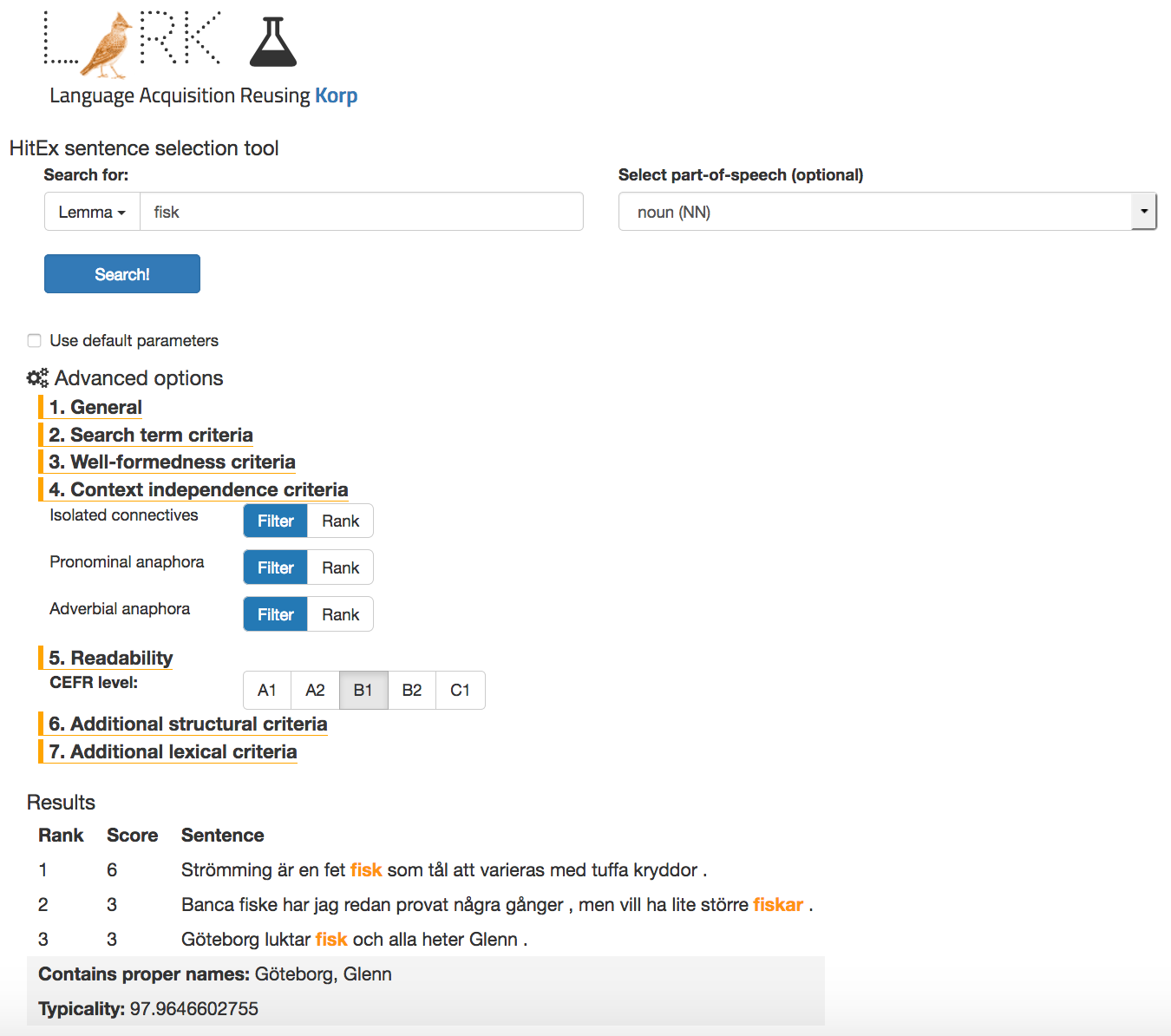}}
\caption{The HitEx user interface with {\normalfont fisk} `fish' as search term. }
\label{img:gui}
\end{figure}

To provide access to our sentence selection algorithm to others, we have integrated it into a freely available learning platform, Lärka.\footnote{\url{https://spraakbanken.gu.se/larkalabb/hitex}.} With the help of a graphical interface, shown in Figure~\ref{img:gui}, users can perform a sentence selection customized to their needs. Under the advanced options menu, users can choose which selection criteria presented in Table~\ref{tab:criteria} to activate as filters or rankers. 
Moreover, the selection algorithm will serve as a seed sentence provider for automatically generated multiple-choice exercises for language learners within the same platform. The sentence selection algorithm is also available as a web service that others can easily integrate in their own systems.

\section{A user-based evaluation}
\label{sec:evaluation}
The main objective when developing our candidate selection algorithm was to identify seed sentences for L2 learning exercises. In absence of an annotated dataset for this task in Swedish, we tested the performance of HitEx with the help of a user-based evaluation.
We assessed the goodness of the candidate sentences in two ways: (i) through L2 teachers confirming their suitability, (ii) by inspecting whether L2 learners' degree of success in solving exercise items constructed based on these candidates matched what is typically expected in L2 teaching. This provided us with information about the extent to which the set of criteria proposed in Section \ref{sec:implementation} was useful for identifying suitable seed sentences. The evaluation sentences and the associated results will be available as a dataset on \url{https://spraakbanken.gu.se/eng/resources}.

\subsection{Participants}
The participants consisted of 5 teachers of L2 Swedish from different institutions and 19 students from a language school targeting young adults newly arrived to Sweden. Participating students were between ages 16 and 19 with a variety of native languages including several Somali and Dari speakers. The proportion of female and male students was approximately equal. The CEFR level of students is assessed on a regular basis with a two-month interval in their school. In our evaluation, as a point of reference for students' CEFR level, we referred to the levels achieved on their latest assessment test. According to this, 3 students were at A1 level, and the remaining 16 were a 50--50\% split between A2 and B1 levels.

\subsection{Material and task}

\label{sssec:material}
To create the evaluation material, we retrieved a set of sentences from Korp for CEFR levels A1--C1 using HitEx. To perform the Korp concordance search, we used lemmas from SVALex whose level corresponded to the level of the sentences we aimed at identifying. We used a lemma-based search and the parts of speech included nouns, verbs and adjectives. The sentences have been selected from 10 different corpora including novels and news texts. For each search lemma, a maximum of 300~matching Korp sentences were fed to the sentence selection algorithm, out of which only the top ranked candidate for each lemma was included in the evaluation material. Most selection criteria were used as filters, but typicality, proper names, KELLY and SVALex frequencies were used as rankers. Modal verbs were allowed in the sentences and the position of the search term was not restricted. Sentence length was set to a minimum of 6 and a maximum of 20 tokens. The threshold used for the percentage of non-alphabetic and non-lemmatized tokens was 30\%.

Teachers received 330 sentences to evaluate, evenly distributed across the 5~CEFR levels A1--C1. The sentences were divided into two subgroups based on their level, at least two teachers rating each sentence. One set consisted of A1--B1 level sentences and the other of sentences within levels B1--C1. (B1 level sentences have been evenly split between the two subsets.) There was a \emph{common subset} of 30 sentences from all 5~CEFR levels which was rated by all 5 teachers.
Besides an overall score per sentence reflecting the performance of the combination of all criteria from Table~\ref{tab:criteria}, we elicited teacher judgements targeting two criteria in particular, which were focal points during the implementation of HitEx, namely context independence and L2 complexity (see Sections~\ref{ssec:context_indep_crit} and \ref{ssec:l2_complexity_crit} respectively).
No specific exercise type needed to be considered for evaluating these aspects, but rather a more application-neutral scenario of a learner reading the sentence.
Teachers rated the three dimensions on a 4-point scale as defined in Table~\ref{table:eval_scale}. Besides these aspects, teachers were also required to suggest an alternative CEFR level if they did not agree with the one predicted by the system. 

\begin{table} 
\centering
\begin{tabular}{cl}
\toprule
\multicolumn{2}{l}{\it \bf The sentence...}\\
\midrule
1 & \it ... doesn't satisfy the criterion. \\
2 & \it ... satisfies the criterion to a smaller extent.\\
3 & \it ... satisfies the criterion to a larger extent.\\
4 & \it ... satisfies the criterion entirely.\\
\bottomrule
\end{tabular}
\caption{\label{table:eval_scale} Evaluation scale. }
\end{table}

To investigate further whether our selection criteria with the chosen setting produced good seed sentence candidates at the CEFR levels predicted by our L2 complexity criteria, we observed 
L2 learners' performance on exercise items created out of these sentences. Exercise generation requires a number of additional steps after the selection of seed sentences, many of which are open research problems.  Therefore, we opted for a semi-automatic approach to the generation of these exercises. We manually controlled the combination of sentences into exercises and the selection of a \textit{distractor}, an incorrect answer option which did not fit into any sentence, in order to reduce potential ambiguity in answer options. A subset of the sentences given to teachers were used as exercise items so that teachers' ratings and students' answers could be correlated.

The exercise type chosen was \emph{word bank}, a type of matching exercise, since this posed less challenges when selecting distractors compared to multiple-choice items. Word bank exercises consist of a list of words followed by a list of sentences, each containing a gap. Learners' task is to identify which word is missing from which sentence. We created worksheets consisting of word bank exercises in Google Forms\footnote{\url{https://docs.google.com/forms/}.}. 
To lower the probability of answering correctly by chance, we added a distractor. Students had to provide their answers in a grid following the list of candidate words and the gapped sentences. The missing word to identify (and its position) corresponded to the search term used to retrieve the sentence from Korp. 
Worksheets consisted of 9~exercises with 5~sentences each, amounting to a total of 45~sentences. (The only exception was A1 level, where students had 2~exercises less.) Students had 60~minutes to work with the exercises, including 5 minutes of instructions. Students worked individually in a computer lab, access to external resources was not allowed. 

The difficulty of the exercises varied along two dimensions: in terms of their CEFR level and in terms of the similarity of the morpho-syntactic form of the candidate words included in the word bank. A worksheet for a certain level contained 5~exercises from the same level as well as 2~exercises from one level below and one level above. In 5~exercises, the word bank consisted of lexical items with the same morpho-syntactic form (e.g.~only plural neuter gender nouns), while 4~exercises had a word bank with mixed POS. The latter group of exercises was somewhat easier, since, besides lexical-semantic knowledge, students could identify the correct solution also based on grammatical clues such as inflectional endings. 

\subsection{Results and discussion}
Below, we present teachers' and students' results on the evaluation material.

\subsubsection{Teachers}
To understand to which extent our set of criteria was able to select suitable seed sentences overall as well as specifically in terms of L2 complexity and context independence, we computed average values and standard deviation (\textsc{StDev}) over L2 teachers' ratings. (8 sentences had to be excluded between A1-B1 levels due to missing values.) The results are presented in Table~\ref{table:teacher_crit_res}. 

\begin{table} 
\centering
\begin{tabular}{cccc}
\toprule
\bf Criterion & \bf  \# of raters & \bf  Average & \bf \textsc{StDev} \\
\midrule
L2 complexity & 5 & 3.18 & 0.53\\
Context independence & 5 & 3.05 & 0.56\\
Overall suitable (all criteria) & 4 & 3.23 & 0.73\\
\bottomrule
\end{tabular}
\caption{\label{table:teacher_crit_res} Average teacher-assigned rating per criteria.}
\end{table}

As for the criterion of context independence, 80\% of the sentences were found suitable (received an average score higher than 2.5), and 61\% of the sentences received score 3 or 4 from at least half of the evaluators.
Besides rating the three dimensions in Table~\ref{table:teacher_crit_res}, teachers also provided an alternative CEFR level in case they did not agree with the CEFR level suggested by the system. HitEx correctly assessed L2 complexity for 64\% of sentences based on teachers' averaged CEFR label, and in 80\% of the cases the system's CEFR level coincided with at least one teacher's decision. Besides comparing system-assigned and teacher-assigned levels, we measured also the inter-rater agreement (IAA) among the teachers. We used \emph{Krippendorff's $\alpha$} measuring observed and expected disagreement, since it is suitable for multiple raters. An $\alpha$ = 1 corresponds to complete agreement, while $\alpha$ = 0 is equivalent to chance agreement. The inter-rater agreement results among teachers are presented in Table~\ref{table:iaa_cefr}. The extent of agreement among teachers was considerably higher than chance agreement, but it still remained below what is commonly considered as reliability threshold in annotation tasks, namely $\alpha$ = 0.8. CEFR level assignment for sentences thus seems to be a hard task even for teaching professionals.

\begin{table} 
\centering
\begin{tabular}{ccccc}
\toprule
\bf SentID & \bf \# sents & \bf \# raters & \bf CEFR & \bf IAA \\
\midrule 
1-38 & 38 & 5 &  A1-C1 & 0.65 \\
39-188 & 142 & 2 & A1-B1 & 0.68 \\
189-338 & 150 & 3 &  B1-C1 & 0.53 \\
\midrule
Tot/Avg & 330 & 5 & A1-C1 & 0.62 \\
\bottomrule
\end{tabular}
\caption{\label{table:iaa_cefr} Inter-rater agreement for CEFR level assignment.}
\end{table}

Besides inter-rater agreement in terms of  $\alpha$, we considered also the distance between the CEFR levels assigned by all teachers compared both to each other and to HitEx (Table \ref{table:teacher_lbl_dist}). This would provide us information about the degree to which teachers' accepted our system's assessment of L2 complexity. CEFR levels were mapped to integer values for this purpose, and \textit{averaged pairwise distances} between the levels have been computed in all cases. 
Surprisingly, teachers agreed with each other exactly on the CEFR level of sentences in only half of the cases, which shows that the exact CEFR level assignment on a 5 point scale is rather difficult even for humans. The percentage of sentences that teachers agreed on with HitEx completely (distance of 0) was slightly (4\%) higher than the extent to which teachers agreed with each other. This may be due to the fact that teachers were confirming the system-assigned CEFR levels, but did not have information about each others' answers. Teacher-assigned CEFR levels remained within 1 level of difference when compared to each other in almost all cases and compared to the system for 92\% of the sentences. 
All in all, the automatic CEFR levels predicted by HitEx were accepted by teachers in the majority of cases within 1 level distance.

\begin{table} 
\centering
\begin{tabular}{ccc}
\toprule
\bf Level &  \bf Teacher -  & \bf Teacher - \\
\bf Distance & \bf Teacher & \bf System I \\
\midrule
0 & 50.0 & \bf 53.9 \\
1 & \bf 49.4 & 37.9 \\
2 & 0.6 & 6.7 \\
$\geq$ 3 & 0.0 & 1.5 \\
\bottomrule
\end{tabular}
\caption{\label{table:teacher_lbl_dist} Percentage of sentences per assigned CEFR label distance.}
\end{table}

Finally, we computed the \textit{Spearman correlation coefficient} for teachers' scores of overall suitability and the two target criteria, L2 complexity and context independence, to gain insight into how strongly associated these two aspects were to seed sentence quality according to our evaluation data. The correlation over all sentences was $\rho=0.34$ for L2 complexity and $\rho=0.53$ for context dependence. The maximum possible value is $\rho=1$ for a positive correlation and $\rho=-1$ for a negative one. Both criteria were thus positively associated with overall suitability: the more understandable and context-independent a sentence was, the more suitable our evaluators found it overall. Out of the two criteria, context dependence showed a somewhat stronger correlation. 

\subsubsection{Students}

First, based on students' responses, we computed \textit{item difficulty} for each exercise item, which corresponds to the percentage of students correctly answering an item, a higher value thus indicating an easier item \cite{crocker1986introduction}. The average item difficulty over all exercises was 0.62, corresponding to 62\% of students correctly answering items on average. Table~\ref{table:res_exe_type} shows additional average item difficulty scores divided per CEFR level, exercise type (distractors with same or different morpho-syntactic form) and POS. Values were averaged only over the exercise items that were of the same CEFR level as the answering students' level according to the system. 

\begin{table*} 
\centering
\begin{tabular}{ccccccccc}
\toprule
\textbf{Ex. type} & 
\multicolumn{3}{c}{\bf \textsc{Same POS+Infl}} & 
\textbf{Avg} & 
\multicolumn{3}{c}{\bf \textsc{Mixed POS+Infl}} & 
\textbf{Avg} \\
\midrule
\textbf{CEFR} & \textbf{A1} & \textbf{A2} & \textbf{B1} & & \textbf{A1} & \textbf{A2} & \textbf{B1} & \\
\midrule
Noun & 0.67 & 0.83 & 0.73 & \bf 0.74 & 0.67 & 0.52 & 0.65  & \bf 0.61\\
Verb & 0.50 & 0.69 & 0.69 & \bf 0.63 & 0.0 & 0.62 & 0.77 & \bf 0.46 \\
Adjective & - &- & - & - & 0.58 & 0.56 & 0.62 & \bf 0.59\\
\midrule
\bf Avg & 0.59 & 0.76 & 0.71 & \bf 0.69 & 0.42 & 0.57 & 0.68 & \bf 0.55 \\
\midrule
\textbf{Overall} & \multicolumn{8}{c}{\bf 0.62} \\
\bottomrule
\end{tabular}
\caption{\label{table:res_exe_type} Average item difficulty per exercise item category, POS and CEFR level.}
\end{table*}

To be able to measure whether the item difficulty observed in our students' results matched the values one would typically expect in L2 teaching, we calculated the \textit{ideal item difficulty} (IID) score for our exercises, which takes into consideration correct answers based on chance. We used the formula proposed by \citeasnoun{thompson1985using} presented in [\ref{formula:iid}], where $P_C$ is the probability of correct answers by chance. 

\begin{equation}
\label{formula:iid}
IID = P_C + \frac{1-P_C}{2}
\end{equation}

Our exercises consisted of 5 gapped items and 6 answer options in the word bank. Students had, thus, a chance of 1/6 for filling in the first item, 1/5 for the second item etc., which corresponds to an average $P_C$ of $(0.167+0.2+0.25+0.333+0.5)/5=0.29$ for the whole exercise and, consequently, an IID score of 0.645, that is 64.5\% of students correctly answering. The observed item difficulty averaged over all students and exercise items of our evaluation was 62\%, which is only slightly lower than the ideal item difficulty. If we break down this average to students' CEFR levels, we can notice that for A1 students the exercises were considerably more challenging than they should have been according to the ideal threshold. Only 51\% of them responded correctly A1-level exercise items. Our sample size was particularly small, however, at this level, thus further evaluations with additional students would be required to confirm this tendency. A2 and B1 level students produced considerably better results: averaging over exercise types and POS, 66.5\% and 69.5\% of them respectively answered correctly the items of their levels. This indicates that the set of criteria proposed in Section \ref{sec:implementation} can successfully select seed sentences for L2 exercises for students of A2 and B1 levels. 

\begin{figure}
\centering
\includegraphics[trim = 7mm 0mm 0mm 0mm, clip, width=0.90\textwidth]{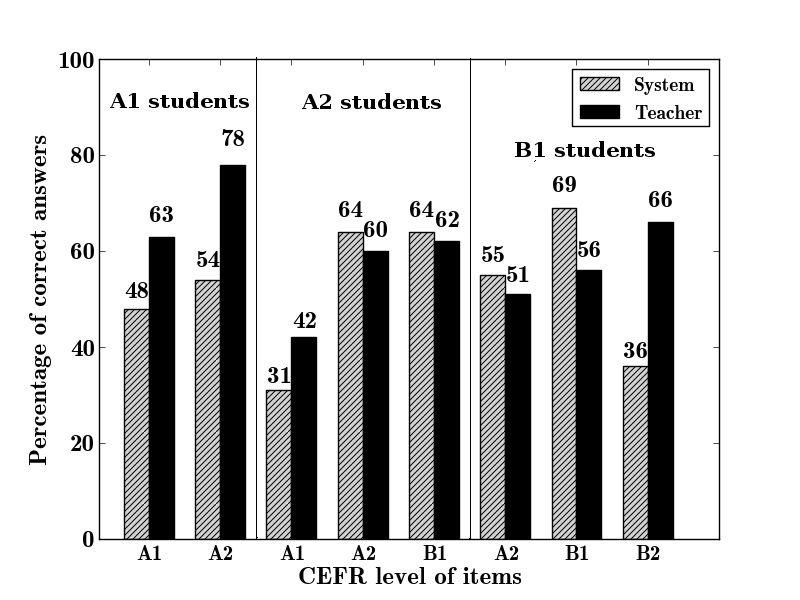}
\caption{Correct answers per averaged teacher and system CEFR level.}
\label{img:stud_perf_S_vs_T_EXT}
\end{figure}

Contrary to what one might expect, exercise items with distractors bearing different morpho-syntactic forms proved to be actually harder for students compared to items with the same POS and inflection based on our evaluation data. The latter would be inherently harder since only lexico-semantic information can contribute to solving the exercises without the help of grammatical clues. As the item difficulty values show in Table~\ref{table:res_exe_type}, approximately 14\% more students answered correctly exercise items with distractors with the same morpho-syntactic tags, an outcome, which, however, may also depend on the inherent difficulty of the sentences presented. As mentioned in Section~\ref{sssec:material}, the work sheets also included exercises constructed with sentences belonging to one CEFR level higher and lower than students' level. This allowed us to further assess whether the CEFR levels suggested based on the L2 complexity criterion were appropriate. We display in Figure~\ref{img:stud_perf_S_vs_T_EXT} students' performance based on the CEFR level of exercise items comparing the system-assigned and the teacher-suggested CEFR levels for the items.

As Figure~\ref{img:stud_perf_S_vs_T_EXT} shows, at A1 level students answered a larger amount of items according to the CEFR level determined by teachers (63\%, vs.~48\% with the system-assigned CEFR). The percentage of correct answers at A2 and B1 levels, however, showed more consistency with the levels assigned by our L2 complexity criterion: 64\% (A2) and 69\% (B1) correct answers based on our system's CEFR levels, vs.~60\% (A2) and 56\% (B1) with teacher-assigned levels. When considering these scores, however, it is worth noting that both teachers and the system were assessing only seed sentence difficulty, not the overall difficulty of the exercises. A few additional aspects play a role in determining the difficulty of exercise items, e.g.~the selected distractors \cite{beinborn2014predicting}, nevertheless the observed tendencies in error rates provide useful insights into the suitability of seed sentences in terms of L2 complexity.

\section{Conclusion}
\label{sec:conclusion}
We presented a comprehensive framework and its implementation for selecting sentences useful in the L2 learning context. The framework, among others, includes the assessment of L2 complexity in sentences and their independence of the surrounding context, both of which are relevant for a wide range of application scenarios. To our knowledge, this is the first comprehensive study addressing automatic seed sentence selection for L2 learning exercises. We invested considerable effort into creating a system that would yield pedagogically more relevant results.
We conducted an empirical evaluation with L2 teachers and learners to gain insights into how successfully the proposed framework can be used for the identification of seed sentences for L2 exercises. Although the sample size was somewhat limited, the evaluation yielded very promising results. On average, the selected sentences lived up to teachers' expectations on L2 complexity, context independence and overall suitability. 
The exercises constructed with the use of the selected sentences were overall somewhat hard for beginners, but they were of an appropriate difficulty level for learners at subsequent stages.
Moreover, learners' error rates at some levels correlated even slightly better with the CEFR levels predicted by our system than the averaged levels assigned by teachers.  
All in all, the evaluation indicated that the described system has good potentials to enhance language instruction by either assisting teaching professional when creating practice material or by providing self-learning opportunities for learners in the form of automatically generated exercises. Although our main focus was on seed sentences selection, the proposed system can be useful also for the identification of dictionary example sentences.  

Future work could include a version of the system aware of word senses, both as search terms and as entries in the word lists applied. This would also enable searching for sentences belonging to specific topics or domains. Moreover, additional information about learners' lexical knowledge could be incorporated, for example, based on learner-written essays. 
Another valuable direction of further research would be the extension of the algorithm to multiple languages, for example through the use of universal POS and dependency tags.
Finally, collecting additional data on how learners perform on the exercises constructed out of the selected sentences could also provide further indication on the quality and usefulness of the proposed algorithm. 


\bibliography{TAL_biblio_ex}

\end{document}